\pdfoutput=1
\documentclass[11pt]{article}

\usepackage{emnlp2022}
\usepackage{booktabs}

\usepackage{url}
\usepackage{times}
\usepackage{latexsym}
\usepackage{graphicx}
\usepackage{diagbox}
\usepackage{wrapfig}

\usepackage{amsmath,amsfonts,bm}

\def\eqref#1{equation~\ref{#1}}
\def\1{\bm{1}}

\DeclareMathAlphabet{\mathsfit}{\encodingdefault}{\sfdefault}{m}{sl}
\SetMathAlphabet{\mathsfit}{bold}{\encodingdefault}{\sfdefault}{bx}{n}

\def\gD{{\mathcal{D}}}

\usepackage[T1]{fontenc}
\usepackage[utf8]{inputenc}

\usepackage{microtype}
\usepackage{xcolor}
\usepackage{xspace}

\newcommand{\dataset}{\textsc{M2D2}\xspace} %teal}{{\normalsize [#1 --MR]}}}
\newcommand{\numtokens}{8.5B\xspace} %teal}{{\normalsize [#1 --MR]}}}

\title{\dataset: A Massively Multi-Domain Language Modeling Dataset}

\author{Machel Reid$^{1}$\thanks{\ \ Currently at Google Research}, Victor Zhong$^2$, Suchin Gururangan$^2$, Luke Zettlemoyer$^2$\\
$^1$The University of Tokyo, $^2$University of Washington\\
\texttt{machelreid@google.com}, \texttt{\{vzhong,sg01,lsz\}@cs.washington.edu}}

\begin{document}
\maketitle
\begin{abstract}

    We present \dataset, a fine-grained, massively multi-domain corpus for studying domain adaptation in language models (LMs). \dataset consists of \numtokens tokens and spans 145 domains extracted from Wikipedia and Semantic Scholar. Using ontologies derived from Wikipedia and ArXiv categories, we organize the domains in each data source into 22 groups. This two-level hierarchy enables the study of relationships between domains and their effects on in- and out-of-domain performance after adaptation. We also present a number of insights into the nature of effective domain adaptation in LMs, as examples of the new types of studies \dataset enables. To improve in-domain performance, we show the benefits of adapting the LM along a domain hierarchy; adapting to smaller amounts of fine-grained domain-specific data can lead to larger in-domain performance gains than larger amounts of weakly relevant data. We further demonstrate a trade-off between in-domain specialization and out-of-domain generalization within and across ontologies, as well as a strong correlation between out-of-domain performance and lexical overlap between domains.\footnote{We release our dataset publicly  at \url{https://github.com/machelreid/m2d2}.}
    
% include public models for camera ready
% 	Language models are typically trained on large-scale, hetereogeneous corpora, containing many domains to improve coverage and performance. However, despite the importance of data-diversity and domains in NLP, there has been no fine-grained study on transfer between domains in a fine-grained massively multi-domain setting. To this end, we present \dataset, the first fine-grained massively multi-domain language modeling corpus, consisting of over 100 domains organized hierarchically extracted from both Wikipedia and Semantic Scholar. \dataset includes over 100 domains, gathered from scientific research papers and encyclopedic text, organized into 22 broader domains using human-informed onotologies with over 5B tokens. We use this dataset, and perform initial experiments and study \textit{which features that are important for transfer} and \textit{what is transferred during domain adaptation in different settings?}\footnote{\dataset will be open-sourced upon acceptance.}
   % \victor{add experiment findings}
\end{abstract}

\section{Introduction}

% \begin{figure*}[th]
% \includegraphics[trim={1cm 2cm 1cm 1cm}, width=\linewidth]{wikidomains.pdf}
% \caption{Visualization of the hierarchies contained within \dataset. Data derived from Wikipedia is shown on the left, while data from Semantic Scholar is shown on the right. \mr{add semantic scholar data}}
% \end{figure*}

% Pre-trained language models, language models pre-trained on large-scale text corpora, provide the back bone for much of NLP done today. 
% Understanding the sub-groups, or \textit{domains}, contained within large swaths of text corpora is crucial to developing effective language systems. 

% \suchin{I sort of think we could start the paper out instead by saying that language is highly hetereogeonous, but treated monolithically. Some works have started to think about broader domains, but we're  proposing to better characterize language variation with a massively multi-domain dataset.}

Even though they can contain a wide variety of different types of domains, the texts that make up the corpora used to train and evaluate language models (LMs) are often treated as if they are all the same. This makes it challenging to characterize LM performance under diverse data distributions and understand how to effectively adapt LMs to new ones. To address these challenges, we develop \dataset, %\footnote{\textbf{M}assively \textbf{M}ulti-\textbf{D}omain language modeling \textbf{D}ataset} 
a \textbf{M}assively \textbf{M}ulti-\textbf{D}omain \textbf{D}ataset, with 145 subdomains and a human-curated hierarchy for studying fine-grained domain adaptation.

\begin{figure}[t]
\includegraphics[width=\linewidth]{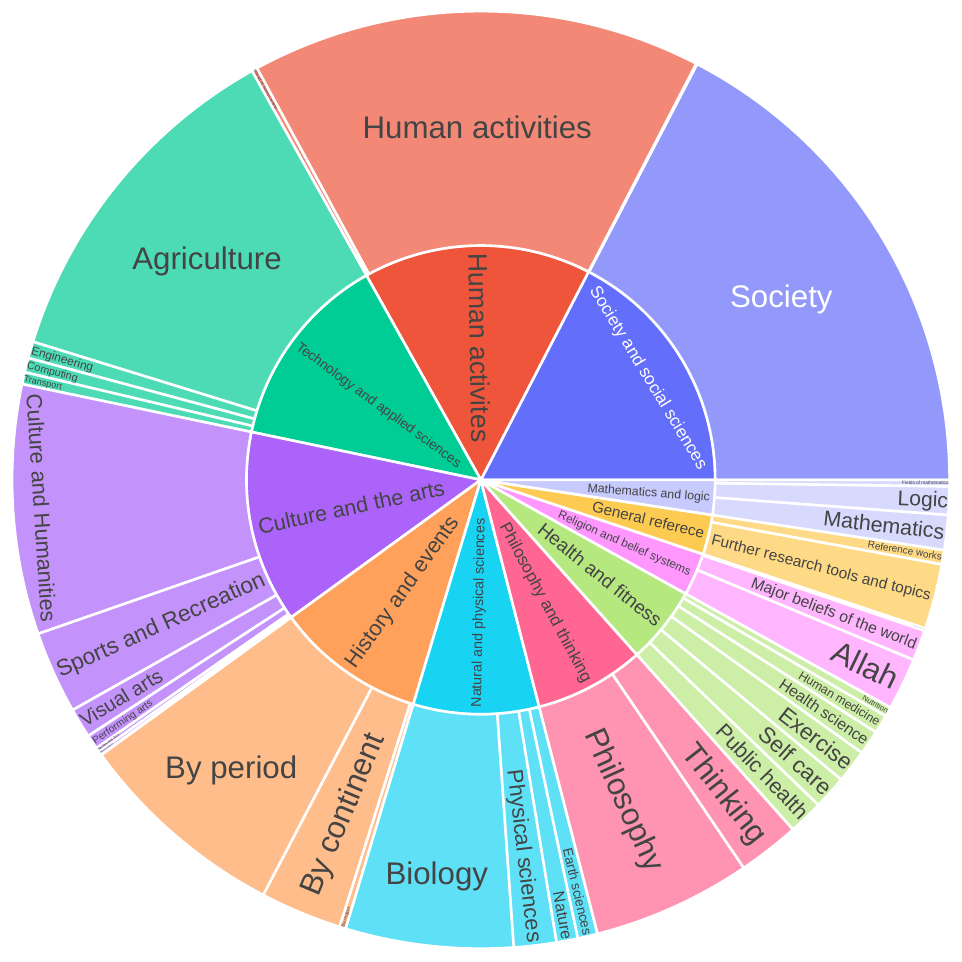}
\caption{Visualization of the two-level fine-grained domain hierarchy in the Wikipedia portion of \dataset.}% \victor{refer to this figure from intro and elsewhere.}}
\label{fig:wiki}
\end{figure}

\begin{figure}[t]
\includegraphics[width=\linewidth]{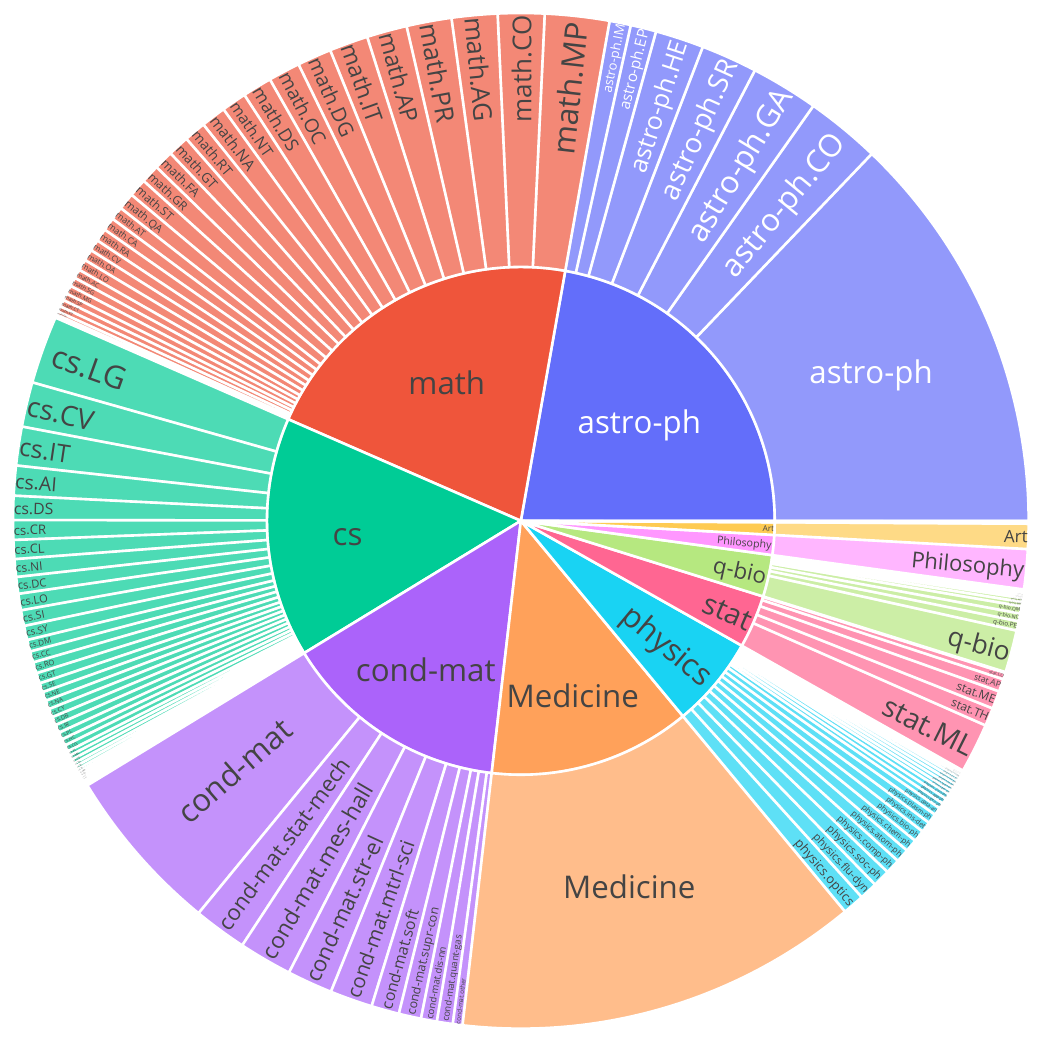}
\caption{Visualization of the hierarchies contained within the S2ORC portion of \dataset.}
\label{fig:s2orc}
\end{figure}

% \mr{justifying dataset by mentioning brief contextualization with prev. work}
% \dataset follows a recent line of work on domain adaptation of large language models \citep{gururangan2020dont,gpt-j,brown2020language,dery2021should}.
Prior work on domain transfer focuses on a small number of broad domains (typically 4-20; \citealp{gururangan2020dont,gao2021pile,gururangan2021demix}).
%However naturally occurring text can contain domains at a much finer granularity, which are often closely related if not perfectly the same. 
% \mr{what is good about \dataset; massive, fine-grained, human-informed, hierarchical}
%\dataset provides a curated 40GB language modeling corpus for studying domain adaptation in language models.
%To better study such phenomena, 
In contrast, domains in \dataset are fine-grained and organized into a hierarchy derived from human-curated ontologies in Wikipedia (Figure~\ref{fig:wiki}) and Semantic Scholar (Figure~\ref{fig:s2orc}).
Unlike prior work, the fine granularity of \dataset enables the study of transfer to naturally occurring data-scarce domains recognized by human curators~(e.g. Philosophy, Public Health, Transport).
This hierarchy enables the study of domain transfer at varying levels of topic granularity.
For instance, how should we combine widely available internet text (entire corpus), text on computer science (coarse domain), and scarce corpus on machine learning (fine domain) to improve performance in the machine learning domain?
% The hierarchical organization of our dataset allows for studies to focus on varying degrees of fine-grainedness, as well as study techniques that take into account hierarchy \citep{gururangan2020dont,Chronopoulou}.
To the best of our knowledge, \dataset is the first dataset that combines fine domain granularity with human-curated domain hierarchy in a massively multi-domain setting.

Using \dataset, we investigate the following questions, as examples of the broad classes of new questions that can be asked:
(1) how well do coarse and fine domains transfer to each other across the hierarchy?
(2) which features and aspects of a domain are important for transfer?
(3) how important is domain specificity versus breadth? 
% And (4) how does model scale influence domain transfer?
We perform preliminary experiments analyzing transfer between similar domains, disparate domains, and hierarchically related domains.
Moreover, we explore how to select source domains to improve transfer performance. %, using features such as noun overlap.

We present baseline experiments using a GPT2~\citep{radford2019language} language model.
We find that (1) more specific data is often more important for performance than larger, less-specific data, shown by our comparison of coarse-grained, fine-grained and coarse-to-fine adaptation comparison (in which coarse-to-fine performed best)
%\victor{what about coarse-to-fine? isn't that the best?}
, (2) vocabulary overlap is a surprising good indicator for transfer, and (3) data source provenance information
%~\victor{I don't know what this means} \mr{the source of the data}
is often a better predictor than ontology when predicting transferability, perhaps indicating a more multi-faceted definition of \emph{domain} could be developed in future work.

% \mr{victor: finally, try to have a forward-looking statement on how we hope people will use this dataset}
% \mr{forward looking statement}
Given the importance of fine granularity domains in language modeling, we hope that \dataset will encourage the community to further study domain transfer: how do we identify hierarchical fine-grained domains in naturally occurring text, and how do we leverage this fine-grained domain hierarchy to improve domain transfer.
\begin{table*}[t]
    \centering
    \resizebox{\textwidth}{!}{\begin{tabular}{llcll}
	\toprule
\bf L1 (Abbrv) & \bf Size & \bf \#L2 & \bf \#Tokens & \bf Examples of L2 domains \\
	\midrule
	Health and fitness $_\text{(HEAL)}$              & 761.2MB & 7    & 116M           &  Exercise, Health Science\\
	History and events $_\text{(HIST)}$              & 1.4GB   & 4    & 226M          & Regions, Periods \\
	Society and social sciences $_\text{(SOCI)}$     & 2.3GB   & 3    & 379M          & Society, social sciences \\
	Technology and applied sciences $_\text{(TECH)}$ & 1.9GB   & 5    & 297M         & Agriculture, Computing  \\
	Culture and the arts $_\text{(CULT)}$            & 2.0GB   & 8    & 289M        & Games and Toys, The arts and entertainment   \\
	Natural and physical sciences $_\text{(NATU)}$   & 1.2GB   & 5   & 189M        & Physical sciences, Earth sciences    \\
	Human activites $_\text{(HUMA)}$                 & 2.1GB   & 3   & 343M        & Impact of human activity    \\
	Mathematics and logic $_\text{(MATH)}$          & 332.3MB & 4   & 52M       & Mathematics, Logic     \\
	General reference $_\text{(GENE)}$                & 385.3MB & 3     & 60M     & Research tools and topics, Reference works     \\
	Religion and belief systems $_\text{(RELI)}$     & 428.0MB & 4    & 64M &       Major beliefs of the world, Belief systems     \\
	Philosophy and thinking $_\text{(PHIL)}$         & 1.0GB   & 3    & 165M &      Philosophy, Thinking      \\

	\midrule
	Mathematics $_\text{(math)}$ & 4.5GB& 26 & 1.4B & Topology, Number Theory \\
	Quantitative Biology$_\text{(q-bio)}$ &1.9GB& 3 &336M& Biomolecules, Cell Behavior\\
	Physics &4.1GB &12 & 737M& General Physics, Biological Physics\\
	Nonlinear Sciences $_\text{(nlin)}$ & 730MB & 5 & 134M & Self-Organizing Systems, Chaotic Dynamics \\
	Condensed Matter $_\text{(cm)}$ & 3.8GB& 10 &688M& Materials Science, Quantum Gases \\
	Economics $_\text{(econ)}$ &67MB&3 & 11M & Econometrics, General Econometrics, Theory\\
	Computer Science $_\text{(cs)}$ &4.5GB& 23 &1.1B& Machine Learning, Databases, Graphics \\
	Statistics $_\text{(stat)}$ & 2.4GB & 4 & 450M & Applications, Methodology \\
	Astrophysics $_\text{(astro-ph)}$ &4.0GB&7 &728M& Earth/Planetary, Cosmology \\
	% Medicine$^\dagger$ &3.6GB& 1 & 567M& --- \\
	Art$^\dagger$ & 575MB&1&98M& --- \\
	Philosophy$^\dagger_\text{(phil)}$ &919MB&1&156M& ---\\
	 \midrule
	\bf Average$_{\pm\text{s.d.}}$ & 1.9G$_{\pm\text{1.7G}}$ & 6.6$_{\pm\text{6.2}}$ & 373M$_{\pm\text{347M}}$ & ---\\
	\bf Total & 41GB & 145 & 8.5B & ---\\
	\bottomrule
    \end{tabular}}
    \caption{Dataset statistics for \dataset. We list L1 domains, with their corresponding sizes, number of L2 domains, number of tokens, and examples of L2 domains. $^\dagger$These domains did not have any subdomains in the arXiv ontology.}
    \label{tab:dataset_statistics}
\end{table*}
\section{\dataset}
% In order to systematically study the properties listed in our problem setup, we created \dataset. In this section, we proceed to describe the dataset collection process, its unique properties, as well as some analysis in order to describe the data.
\dataset consists of a large quantity of fine-grain domains. Unlike prior work that defines the domain of a corpus using its source (e.g.~the web text domain;~\citealp{Chronopoulou}), we derive domains from a human-curated Wikipedia and arXiv ontologies.
%This allows us to build a hierarchy of fine-grained domains recognized by humans without using heuristic definitions.
%The fine-grained and hierarchical nature of \dataset allows us to answer questions regarding which properties of the data result in higher transfer performance. For example, how do we balance the specificity of the source domain with the availability of a source corpus? Given a fine-grained target domain, what fine-grained source domains result in better transfer? Can we automatically identify more performant source domains without training? 
In this section, we describe how \dataset is collected and organized.

\subsection{Domain Organization}
One of the unique properties of \dataset is its hierarchical nature, enabling the study of transfer at different levels of domain granularity.
We assume a particular corpus to have $L_0, \dots, L_K$ levels of hierarchy, where $L_0$ refers to the lowest or most coarse-grained/broad level (i.e.~the whole dataset), and $L_K$ refers to the highest or most fine-grained/specific level. 
A given level of hierarchy $L_i$ contains $N_i$ domains $\gD^i_{N_i}$, 
\begin{equation}
L_i =[\gD^i_0,\dots,\gD^i_j, \dots, \gD^i_{N_i}]
\end{equation}
$\gD^i_j$ is composed of multiple subdomains $\{\gD^{i+1}_0, \dots, \gD^{i+1}_{N_{i+1}}\}$, which are represented in the next level of the hierarchy $L_{i+1}$.
Similarly, we assume that a given subdomain is contained within a larger domain.
% That is, when we directly compare a domain in a lower level of the hierarchy and one of its subdomains at a higher level of the hierarchy, it is certain that the subdomain is contained within the given domain, albeit with multiple other subdomains. 

For the rest of the paper, we use L1 and L2 to represent the two levels of a $K$ level hierarchy that we consider in this paper.

\subsection{Dataset Collection}
We collect \dataset from two resources, Wikipedia and Semantic Scholar. This allows us to explore domain adaptation in a massively multi-domain setting among domains of varying granularity, while also allowing us to test whether our findings hold across different data sources.
\paragraph{Semantic Scholar} We use the S2ORC corpus \citep{lo-etal-2020-s2orc}, a large corpus of English academic papers annotated with extensive metadata. Using this corpus, which is already categorized into L1-domains representing broader fields of academic research (e.g.~Computer Science, Physics), we extract L2-domains by finding a given paper's respective arXiv\footnote{\url{https://arxiv.org}} category (e.g. ``Computation and Language'' $\in$ Computer Science).

\paragraph{Wikipedia} We crawl the Wikipedia ontology,\footnote{\url{https://en.wikipedia.org/wiki/Wikipedia:Contents/Categories}} which lists major categories contained within Wikipedia.
Within these major categories or L1-domains, we then proceed to look up the category pages within a given L1-domain, and gather respective L2-domains.
This procedure yields a hierarchy of domains contained within Wikipedia.
We then download the Wikipedia data dump, which we clean using \texttt{wikiextractor}\footnote{\url{https://github.com/attardi/wikiextractor}} and assign a given page to its respective domain.

\subsection{Unique Properties}
\dataset has the following major unique properties when compared to previous domain adaptation datasets.
First, it is massively multi-domain: we have 145 L2 domains grouped into 22 L1 domains, which allows us to test domain adaptation for language modeling on a variety of axes (such as hierarchy, subject matter, and ontology) that would be more difficult with more coarse-grained datasets.
Second, \dataset is hierarchical: this allows us to also test the performance of domain specificity versus domain breadth in more flexible adaptation settings. 

We describe dataset statistics in Table~\ref{tab:dataset_statistics}, including dataset size (measured in MB/GB), token count (measured by whitespace tokenization), and the number of L2 domains within each L1 domain. \dataset contains \numtokens tokens, with an average of 373 million tokens per L1 domain. Demonstrating the hierarchical nature of \dataset, we also list examples of L2 domains contained within the L1 domains (e.g. Computing $\in$ Technology and Applied Sciences, Topology $\in$ Mathematics) which are also graphically shown in Figures~\ref{fig:wiki} and \ref{fig:s2orc}).
% \suchin{add reference to Table 1, and describe broad statistics - num tokens, size, etc.}
% 
% \suchin{Maybe we should display (or write about) an example of a couple L1 domains from each ontology and their L2s? It's interesting that there are many L2 domains for each L1, might be good to give intuition about why this would exist. }

\subsection{Dataset Splits}
We split each domain into the respective train, validation, and test sets. To prevent data leakage between the domains when pages belong to two or more domains, we construct validation and test sets from pages that are not contained within any other domains on the same level of hierarchy. For example, the page for ``Biotechnology'' overlaps in domain with both \textit{Biology $\in$ Natural and Physical Sciences} and \textit{Engineering $\in$ Technology and Applied Sciences} so this would not be included in any evaluation set due to the potential for direct leakage. However, the page for ``Computer'' is only in \textit{Computing $\in$ Technology and Applied Sciences} and therefore could be included in an evaluation set. We include at least 1 million tokens in the validation and test sets, respectively.
This enables us to have a precise evaluation set of texts that only belong to a single fine-grained domain.

\begin{table*}[t]
    \centering
    \resizebox{\textwidth}{!}{
    \begin{tabular}{lllllllllllllllllllllllllllllllllllll}
	    \toprule
	    \bf Wiki  &  \bf HEAL   &  \bf HIST   &  \bf SOCI   &  \bf TECH   &  \bf CULT  &  \bf HUMA   &  \bf MATH   &  \bf GENE   &  \bf RELI   &  \bf PHIL   &  \bf NATU & Avg \\
	    \midrule
	    GPT2 & 23.1 & 27.5 & 24.5 & 27.8 & 27.5 & 28.9 & 26.6 & 25.9 & 26.3 & 26.2 & 26.7 & 26.5 \\
	    \midrule 
	    L1 & 18.1$_{2.5}$ & 20.9$_{0.5}$ & 19.7$_{0.8}$ & 22.3$_{0.8}$ & 21.2$_{2.3}$ & 23.0$_{1.4}$ & 18.3$_{5.2}$ & 21.6$_{0.8}$ & 19.8$_{0.5}$ & 21.8$_{0.6}$ & 20.8$_{3.2}$ & 20.7 \\ 
	    L2 & 17.5$_{2.7}$ & 17.8$_{1.9}$ & 17.5$_{0.7}$ & 21.8$_{1.0}$ & 21.7$_{2.6}$ & 22.4$_{0.9}$ & 17.8$_{5.2}$ & 20.8$_{1.0}$ & 18.3$_{0.4}$ & 21.0$_{0.4}$ & 21.7$_{1.6}$ & 19.8 \\
	    L1-to-L2 &  \bf 16.8$_{2.7}$ & \bf  16.7$_{2.1}$ & \bf  15.4$_{0.5}$ & \bf 21.4$_{0.9}$ & \bf 20.6 $_{2.6}$ & \bf 22.0 $_{0.8}$ & \bf 17.1$_{5.0}$ & \bf 19.6$_{1.1}$ & \bf 16.9 $_{0.5}$ & \bf 20.5 $_{0.4}$ & \bf 20.3$_{1.5}$ & 18.8\\
	    \midrule
	    \midrule
	    \bf S2ORC  &   \bf Math & \bf Econ &   \bf CS &  \bf CM &  \bf Physics &  \bf Art &  \bf Phil &  \bf Stat &  \bf Q-Bio &  \bf Nlin &  \bf Astro-Ph & Avg \\
	    \midrule
	    GPT2 &  26.1$_{2.8}$ & 28.2$_{2.7}$ & 26.8$_{2.9}$ & 29.7$_{1.2}$ & 32.7$_{2.1}$ & 35.1$_{0.0}$ & 32.9$_{0.0}$ & 22.7$_{7.3}$ & 30.1$_{1.3}$ & 25.5$_{1.4}$ & 31.6$_{1.5}$ & 29.2 \\ \midrule
        L1 & 9.2$_{3.4}$ & 15.9$_{2.2}$ & 15.4$_{4.0}$ & 12.5$_{1.0}$ & 17.1$_{1.7}$ & 27.7$_{0.0}$ & 24.4$_{0.0}$ & 11.0$_{3.5}$ & 22.6$_{2.2}$ & 9.8$_{2.4}$ & 15.5$_{3.2}$ & 16.5 \\
	    L2 & 8.0$_{3.2}$ & 13.4$_{2.1}$ & 15.1$_{6.7}$ & 12.0$_{1.3}$ & 16.5$_{1.3}$ & 27.7$_{0.0}$ & 24.4$_{0.0}$ & 10.2$_{2.5}$ & 21.0$_{1.3}$ & 9.6$_{2.1}$ & 14.0$_{2.3}$ & 15.7 \\
	    L1-to-L2 & \bf 7.5$_{3.2}$ & \bf 12.5$_{2.2}$ & \bf 14.0$_{5.9}$ & \bf 11.5$_{1.0}$ & \bf 16.1$_{1.6}$ & \bf 27.7$_{0.0}$ & \bf 24.4$_{0.0}$ & \bf 9.3$_{3.3}$ & \bf 20.3$_{1.0}$ & \bf 9.2$_{2.1}$ & \bf 12.9$_{2.3}$ & \bf 15.0 \\
	    \bottomrule
    \end{tabular}}
    \caption{In-domain test perplexities, aggregated to each L1 domain. We look at the impact of L1 vs L2 vs L1-to-L2 finetuning settings when compared to simply finetuning on L1. L2 Adaptation is usually more effective than L1 Adaptation, emphasizing the importance of fine-grained domains, with a coarse-to-fine setup using L1-to-L2 Adaptation is most effective. This finding is statistically significant ($p<0.05$; measured using the Kolmogorov-Smirnov test).}
    \label{tab:specificity}
\end{table*}

\begin{table*}[t]
    \centering
    \resizebox{\textwidth}{!}{
    \begin{tabular}{lllllllllllllllllllllllllllllllllllll}
	    \toprule
	    \bf Wiki  &  \bf HEAL   &  \bf HIST   &  \bf SOCI   &  \bf TECH   &  \bf CULT  &  \bf HUMA   &  \bf MATH   &  \bf GENE   &  \bf RELI   &  \bf PHIL   &  \bf NATU & Avg \\
	    \midrule
	    \midrule 
	    L1 & 23.6$_{3.5}$ & 23.2$_{2.0}$ & 22.4$_{2.2}$ & 22.4$_{2.3}$ & 22.3$_{2.2}$ & 22.7$_{2.0}$ & 25.1$_{3.4}$ & 24.2$_{2.3}$ & 24.7$_{2.8}$ & 23.6$_{2.7}$ & 23.3$_{3.2}$ & 23.3\\ 
	    L2 & 26.1$_{3.8}$ & 26.1$_{3.9}$ & 25.7$_{2.7}$ & 26.1$_{3.5}$ & 27.0$_{3.7}$ & 25.6$_{3.6}$ & 28.9$_{6.9}$ & 25.1$_{2.4}$ & 26.3$_{2.9}$ & 24.1$_{2.6}$ & 26.3$_{3.7}$ & 26.1 \\
	    L1-to-L2 & 25.5$_{3.8}$ & 25.9$_{3.8}$ & 25.2$_{2.6}$ & 26.0$_{3.3}$ & 27.0$_{3.7}$ & 25.1$_{3.6}$ & 28.5$_{7.0}$ & 24.5$_{2.4}$ & 26.2$_{2.9}$ & 23.2$_{2.6}$ & 25.2$_{3.7}$ & 25.7\\
	    \midrule
	    \midrule
	    \bf S2ORC  &   \bf Math & \bf Econ &   \bf CS &  \bf CM &  \bf Physics &  \bf Art &  \bf Phil &  \bf Stat &  \bf Q-Bio &  \bf Nlin &  \bf Astro-Ph & Avg \\
	    \midrule
	    L1 &  32.0$_{17.2}$ & 28.8$_{10.9}$ & 23.1$_{10.1}$ & 24.9$_{14.0}$ & 22.8$_{10.6}$ & 26.8$_{3.3}$ & 25.7$_{3.9}$ & 23.4$_{11.5}$ & 23.2$_{11.3}$ & 23.8$_{12.9}$ & 26.2$_{12.8}$ & 25.5 \\
	    L2 & 36.0$_{21.9}$ & 33.4$_{11.1}$ & 32.1$_{18.7}$ & 32.7$_{17.3}$ & 25.4$_{12.1}$ & 26.8$_{3.3}$ & 25.7$_{3.9}$ & 32.7$_{24.7}$ & 33.2$_{19.6}$ & 34.8$_{22.4}$ & 27.2$_{11.4}$ & 30.9 \\
	    L1-to-L2 & 36.8$_{24.8}$ & 31.9$_{12.6}$ & 31.0$_{22.0}$ & 30.2$_{18.2}$ &  24.2$_{11.4}$ & 26.8$_{3.3}$ & 25.7$_{3.9}$  & 30.4$_{23.0}$ & 32.1$_{23.4}$ & 36.5$_{30.8}$ & 27.5$_{15.1}$ & 30.3\\
	    \bottomrule
    \end{tabular}}
    \caption{Out-of-domain test perplexities, aggregated to each L1 domain. We look at the impact of L1 vs L2 vs L1-to-L2 finetuning settings when compared to simply finetuning on L1. We can see that L2 Adaptation and L1-to-L2 Adaptation are generally less performant in out-of-domain settings that L1 Adapted models, given their in-domain specification. The comparison between L1 versus L2 is statistically significant $p<0.01$.}
    \label{tab:ood}
\end{table*}

\begin{table*}[ht]
    \centering
    \resizebox{\textwidth}{!}{
    \begin{tabular}{cccccccccccccccccccccccccccccccc}
	    \toprule
	     \bf Domain &  \bf NATU &  \bf TECH &  \bf SOCI &  \bf HEAL &  \bf HIST &  \bf RELI &  \bf CULT &  \bf GENE &  \bf MATH &  \bf HUMA &  \bf PHIL &  \bf  Avg     \\
	     \midrule
 NATU & --- & 25.5 & 22.1 & 20.0 & 24.5 & 23.5 & 25.7 & 23.7 & 21.0 & 25.6 & 23.2 & 23.3 \\
 TECH & \bf 23.6 & --- & 20.8 & \bf 19.4 & 23.1 & 22.7 & 23.5 & 22.6 & 21.7 & 24.5 & 22.4 & 22.5 \\
 SOCI & 23.8 & 24.2 & --- & 19.8 & 22.3 & 21.7 & \bf 23.4 & \bf 22.0 & 22.6 & 24.1 & \bf 22.0 & \bf 22.4 \\
 HEAL & 24.3 & 25.6 & 21.6 & --- & 24.4 & 23.7 & 25.2 & 24.0 & 25.0 & 26.2 & 23.9 & 23.9 \\
 HIST & 24.7 & 25.3 & 20.7 & 21.8 & --- & 21.4 & 24.2 & 22.8 & 24.0 & \bf 23.9 & 22.6 & 23.0 \\
 RELI & 26.3 & 28.2 & 21.9 & 22.8 & 24.0 & --- & 25.8 & 24.4 & 26.0 & 26.3 & 24.0 & 24.5 \\
 CULT & 23.7 & 24.3 & 20.6 & 20.1 & 23.0 & 22.1 & --- & 22.5 & 22.8 & 24.4 & 22.2 & 22.5 \\
 GENE & 25.4 & 26.4 & 21.8 & 22.1 & 24.2 & 23.2 & 25.4 & --- & 24.5 & 26.2 & 23.3 & 24.1 \\
 MATH & 26.3 & 26.7 & 23.7 & 23.1 & 26.4 & 25.0 & 27.1 & 25.2 & --- & 28.3 & 24.4 & 25.0 \\
 HUMA & 23.9 & \bf 24.0 & \bf 20.1 & 20.7 & \bf 22.0 &\bf  21.3 & 24.1 & 22.3 & 23.2 & --- & 22.3 & 22.5 \\
 PHIL & 25.1 & 25.7 & 21.8 & 21.3 & 24.4 & 22.9 & 24.7 & 23.5 &\bf 20.9 & 26.0 & --- & 23.5 \\
	     \midrule
Avg & 24.4 & 25.3 & 21.3 & \bf 20.8 & 23.6 & 22.5 & 24.6 & 23.1 & 22.7 & 25.3 & 22.9 & 23.3 \\
% \midrule \midrule
%  GPT2 & 26.7 & 27.8 & 24.5 & 23.1 & 27.5 & 26.3 & 27.5 & 25.9 & 26.6 & 28.9 & 26.2 & 26.51 \\
 \bottomrule
    \end{tabular}}
    \caption{Out-of-domain transfer performance between all L1 domains (using abbreviations from Table~\ref{tab:dataset_statistics}) in the Wikipedia portion of \dataset. For each domain, we use the first four letters to refer to itself. The x-axis shows evaluation domains, and the y-axis shows training domains.} % \suchin{either fill out the domain names (avoid abbreviations) or add legend in the caption, don't know what some of these refer to} } %\suchin{are these randomly chosen? If so, indicate that and say we put the rest of the results in appendix? Highlight best performing transfer in each column and average. Also, briefly describe what the takeaway is of this table.}}
    \label{tab:transfer}
\end{table*}
\begin{table*}[ht]
    \centering
    \resizebox{\textwidth}{!}{
    \begin{tabular}{cccccccccccccccccccccccccccccccc}
    \toprule
    \bf Domain & \bf math & \bf econ & \bf cs & \bf cm & \bf physics & \bf Art & \bf Philosophy & \bf stat & \bf q-bio & \bf nlin & \bf astro-ph& \bf Avg \\
    \midrule
math & --- & 25.0 & 22.0 & 21.2 & 35.8 & 66.1 & 57.2 & 19.6 & 38.8 & 13.6 & 43.2 & 32.0 \\
econ & 18.5 & --- & 23.8 & 24.6 & 35.1 & 48.1 & 43.9 & 15.3 & 33.7 & 20.0 & 37.5 & 28.8 \\
cs & 12.6 & 17.6 & --- & 18.0 & 24.5 & 43.2 & 40.1 & 13.9 & 26.6 & 14.0 & 28.6 & 23.1 \\
cm & 13.6 & 20.7 & 21.6 & --- & \bf 17.9 & 55.9 & 50.0 & 16.2 & 26.4 & \bf 13.2 & 25.8 & 24.9 \\
physics & 14.1 & 20.5 & 21.0 & \bf 14.1 & --- & 46.2 & 41.9 & 15.8 & \bf 24.8 & 13.3 & \bf 22.1 & \bf 22.8 \\
Art & 22.9 & 25.8 & 25.9 & 27.5 & 31.1 & --- & \bf 29.0 & 21.3 & 29.1 & 22.6 & 31.7 & 26.8 \\
Philosophy & 20.8 & 24.7 & 23.4 & 26.2 & 31.2 & \bf 30.4 & --- & 20.3 & 28.1 & 21.5 & 31.4 & 25.7 \\
stat & 12.7 & \bf 14.0 & \bf 18.2 & 17.9 & 24.8 & 47.0 & 43.2 & --- & 26.6 & 14.8 & 27.0 & 23.4 \\
q-bio & 13.7 & 18.1 & 19.2 & 14.6 & 20.9 & 48.1 & 42.9 & \bf 13.6 & --- & 14.3 & 26.8 & 23.2 \\
nlin & \bf 11.0 & 19.7 & 20.7 & 13.3 & 22.3 & 51.7 & 45.8 & 15.7 & 25.9 & --- & 25.9 & 23.8 \\
astro-ph & 16.6 & 23.9 & 25.2 & 17.1 & 23.6 & 54.4 & 48.1 & 17.8 & 30.9 & 15.2 & --- & 26.2 \\
\midrule
Avg & \bf 15.1 & 20.5 & 21.5 & 18.8 & 25.8 & 47.1 & 42.4 & 16.4 & 28.5 & 15.7 & 28.7 & 25.5 \\
% GPT2 & 26.1 & 28.2 & 29.8 & 31.1 & 32.7 & 35.1 & 32.9 & 22.7 & 28.3 & 25.5 & 31.6 & 29.5 \\
 \bottomrule
    \end{tabular}}
    \caption{Out-of-domain transfer performance between all L1 domains in the S2ORC portion of \dataset. ``GPT2'' refers to the zero-shot performance of the LM on our dataset. }
    %\suchin{Move this table to appendix; replace this with a table displaying average out of domain performance of all of our adaptation tehcniques; mirroring table 2. }} % \suchin{same comment as for table 2} }
    \label{tab:transfer_s2orc}
\end{table*}
\begin{figure}[t]
	\centering
\includegraphics[width=0.48\textwidth]{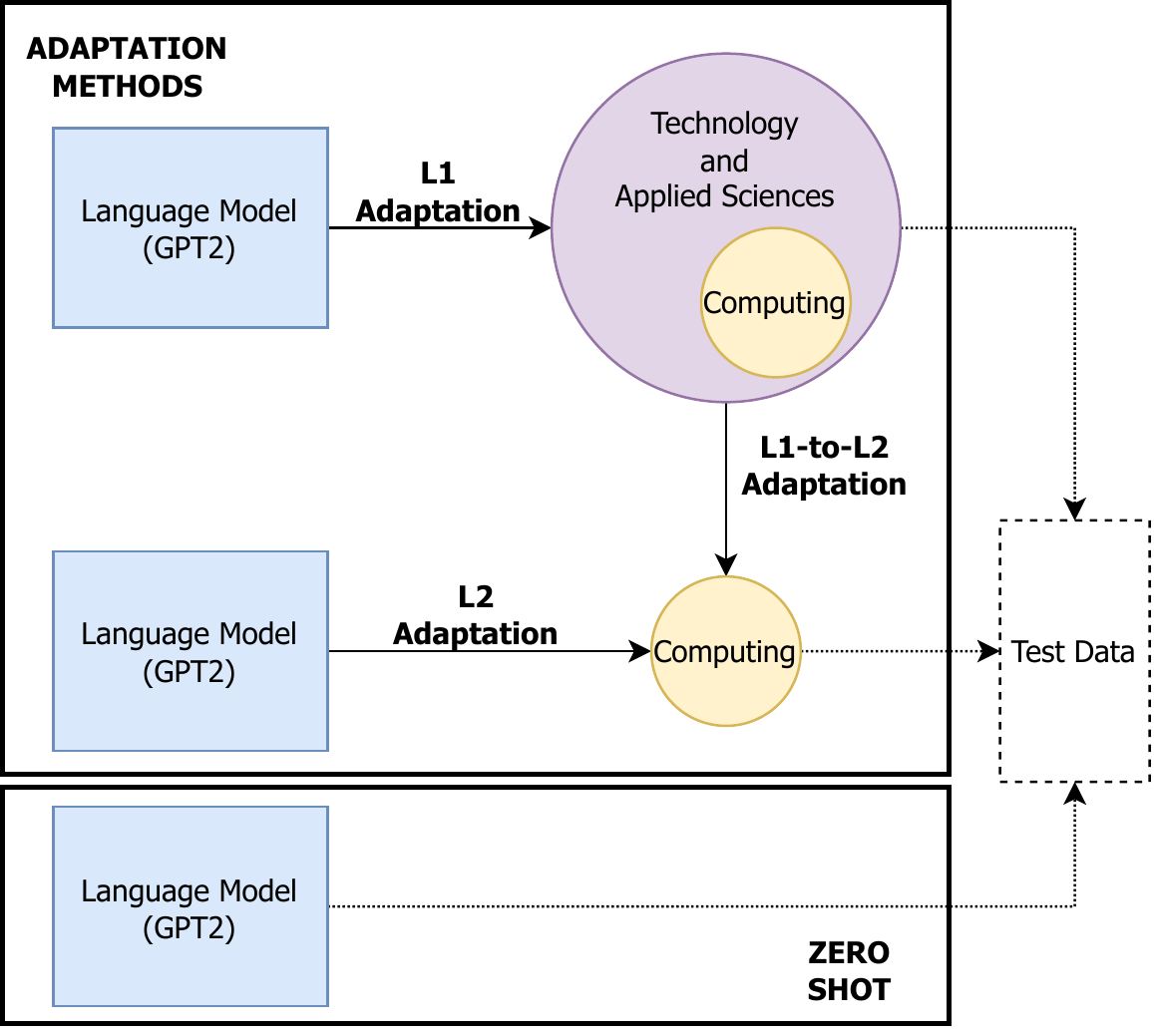}
\caption{The types of domain adaptation that we consider in this work: L1, L2, and L1-to-L2 adaptation. Here, we use ``Technology and Applied Sciences'' to illustrate our L1 domain and ``Computing'' to illustrate our L2 domain. Bold arrows refer to adaptation steps, and dotted lines refer to an evaluation phase. }
\label{fig:adaptation}
\end{figure}

 \begin{table*}[t]
    \small
     \centering
     \resizebox{\textwidth}{!}{\begin{tabular}{lcccccc}
	     \toprule
	     \bf S2ORC  &  \bf  Mathematics  &  \bf Computer Science  &  \bf Art  &  \bf Philosophy  & \bf Physics \\
	     \midrule
	     S2ORC (in-domain) & 9.2 & 15.4 & 27.7 & 24.4 & 17.1 \\
	     Wiki  (in-domain) & 19.6 & 26.8 & 35.3 & 33.4 & 29.6\\
	     S2ORC (out-of-domain) & 15.1 & 21.5 & 47.1 & 42.4 & 25.8 \\
	     \midrule 
	     \midrule 
	     \bf Wiki  &  \bf MATH  &  \bf TECH  &  \bf CULT  &  \bf PHIL & \bf  NATU\\
	     \midrule 
	     Wiki (in-domain) & 18.3 & 22.3 & 21.2 & 21.8  & 20.8  \\
	     S2ORC (in-domain) & 29.6 & 29.5 & 26.8 & 27.0 & 31.5 \\
	     Wiki (out-of-domain) & 22.7 & 25.3 & 24.6 & 22.9  & 22.9\\
	\bottomrule

     \end{tabular}}
     \caption{Transfer performance between corresponding domains(Math$\leftrightarrow$Mathematics and Logic(Math), Computer Science$\leftrightarrow$Technology and Applied Sciences, Art$\leftrightarrow$Culture and the Arts, etc..) in both ontologies. It can be seen that provenance is a stronger indicator of transfer performance on \dataset than ontological correspondence. }
     \label{tab:cross_level}
 \end{table*}

\section{Experiments}
%In our experimental section, we look to test the following properties of \dataset. 

As examples of the types of new studies \dataset enables, we explore a number of key questions about the nature of effective domain adaptation in language models. For example, how does one best specialize a language model to a domain, given an ontology? How well can adapted models be applied out-of-domain, within and across ontologies? What features of target domains are predictive of out-of-domain transfer?

In this section, we present a set of experiments that begin to answer these questions. First, we study the impact of adapting to the L1 and L2 domains of our dataset on  in-domain (\S\ref{sec:adaptation_results}) and out-of-domain (\S\ref{sec:transfer_results}) language modeling performance. Then, we perform an analysis of lexical features in domains that are predictive of out-of-domain performance (\S\ref{sec:lexical_features}). 

% First, we validate whether \dataset captures distinct domains by measuring whether performance on an in-domain test set is consistently improved by our in-domain training.
% We then study the impact of corpora specificity/size by way of various adaptation techniques (detailed in the next subsection).
% In addition, we perform a preliminary study of language model domain adaptation in a massively-multi-domain setting to identify which properties are conducive for transfer (e.g.~noun overlap) and which types of words (e.g entities, pronouns, stopwords) are transferred in different settings (e.g.~transfer between similar domains versus distant domains).

\subsection{Experimental setup}
% \suchin{I think the "scale" question (ie what happens when you increase model size) is pretty interesting, and I think it's an important question we should try to answer if we want LM folks to care about this benchmark. If it's not a lot of work, it would be good to replicate these experiments with one or two larger GPT-2 models; we can probably help you out with these if compute is a concern!}
% (containing 12 transformer decoder \citep{vaswani2017attention} layers, a hidden dimension of 768 and 12 attention heads in each layer) 
In all experiments, we use the 112M GPT2 model \citep{radford2019language} as the baseline model.
Our implementation is based on HuggingFace Transformers \citep{wolf-etal-2020-transformers} and PyTorch \citep{paszke2019pytorch}. All adaptation techniques are performed  using Adam \citep{kingma2014adam}, dropout value of $0.2$ \citep{JMLR:v15:srivastava14a}, using a learning rate of 5e-5 and a batch size of 64000 tokens. We train all models for a maximum of 1 million iterations and perform early stopping over the validation set. All experiments are run on 8 NVIDIA V100 GPUs. 
%\suchin{how long are you training for each setting?} 
% \suchin{add all HPs to appendix if you have time, ask suchin for template}

% We also report out of domain \mr{define this} transfer performance for each micro and macro domain.
% \subsection{Fine-tuning Setups}
\label{sec:ft_settings}
% \suchin{see above comments on nomenclature}

When adapting our GPT2 model to domains in \dataset, we use one of three settings:

\paragraph{L1 Adaptation} We continue training on a given L1 domain (e.g.~Computer Science). 
% In addition to measuring in-domain performance, we also measure performance on other L1-domains as well as its L2-domains. 
\paragraph{L2 Adaptation} We continue training on a given L2 domain (e.g.~Machine Learning). 

% We evaluate on L2 domains within its corresponding L1 domain. % \suchin{are you training the same amount of time as L1?}

% In these settings, as dataset size tends to be smaller, we find that the model converges comparatively quickly when compared to L1 Adaptation.
\paragraph{L1-to-L2 Adaptation} Given a L2 domain (e.g.~Machine Learning), we first perform L1 adaptation on its corresponding L1 domain (e.g.~Computer Science), and then we further perform L2 adaptation. This setting similar to multi-stage adaptive pretraining approaches used for supervised tasks \citep{gururangan2020dont}. % \suchin{are you training for the same amount of time as the other baselines?}

For all techniques, we evaluate test perplexity on L2 domains validation sets.
Due to the large quantity of L2 domains, we aggregate L2 results by their corresponding L1.
For each ontology, we report the average and standard deviation (average$_\text{s.d.}$) of perplexities across L2 domains in each L1. 
% \suchin{The findings here are a bit all over the place, maybe we can consolidate them as answers to $k$ (3-4) research questions, similar to how we did it here: https://arxiv.org/pdf/2111.07408.pdf, and we can preface these questions in the beginning of the document. As I see it, we have 3-4 questions around 1) transfer within/between domains and ontologies 2) domain-specificity vs breadth  3) what features of a domain correlate with transfer 4) maybe something about model size and/or pretraining data?  }  

% In this section, we detail the experiments we perform on this data. Particularly, we aim to analyse aspects of transfer performance that have been previously overlooked, while also seeing whether previous findings also hold on our data.

\subsection{In-Domain Results}
\label{sec:adaptation_results}

The first set of experiments in this study considers the impact of adapting the language model to different levels of the \dataset ontologies. We only consider in-domain perplexity, or the perplexity of model on the domain it is adapted to.

% We are interested in analyzing the specialization of language models to each domain of our dataset, by computing in-domain perplexity, or the perplexity of the language model it was adapted to.

\paragraph{Adaptation improves in-domain performance despite pretraining.}
Table \ref{tab:specificity} shows test-set perplexities on L2 domains, averaged across each L1 domain, after performing each adaptation technique (see Appendix on full results). First, we observe that all proposed adaptation techniques improve performance over the base GPT-2 model. This highlights the effectiveness of adaptation in improving in-domain performance, even when considering domains that the language model has likely been exposed to during pretraining (as is the case with Wikipedia; L1 adaptation results in a 5.8 decrease in perplexity). For domains which the language model is less likely to have been exposed to during pretraining, this is more pronounced (as is the case with S2ORC; L1 adaptation results in a 12.7 decease in perplexity). 

\paragraph{Specificity and hierarchy is more important than broad coverage in adaptation.}
Next, we observe that in most cases, adapting to L2 domains is more beneficial to in-domain performance than adapting to L1 domains. Adaptation to finer-grained domains better specializes a language model, even though these domains are much smaller than their L1 counterparts. Finally, we observe that using L1-to-L2 adaptation further benefits in-domain performance over L2 adaptation in all cases. 
Our results suggest that adapting to smaller amounts of domain-specific data leads to more effective in-domain specialization than adapting to large quantities of data that may be more weakly domain-relevant. Moreover, the best results may be achieved by organizing the target domain into subsets of broader and fine-grained data, and adapting along this hierarchy. However, this approach has increased memory and computational requirements relative to solely relying on L1 Adaptation.

\subsection{Out-of-Domain Results}
\label{sec:transfer_results}

We also study the effects of our adaptation techniques on out-of-domain performance, by performing zero-shot inference with adapted models on domains (e.g.~Art) \emph{other} than the ones they are adapted to (e.g.~Machine Learning). We first transfer models between domains in the same ontology (e.g.~Wikipedia $\rightarrow$ Wikipedia), and then across ontologies (e.g.~Wikipedia $\rightarrow$ S2ORC). 

\paragraph{L2 Adaptation decreases out-of-domain performance.} We show out-of-domain performance for each adaptation technique in Table~\ref{tab:ood}. We show that conversely to L2 and L1-to-L2 adaptation which significantly improved in-domain performance, this comes with the tradeoff at less performance in both L2 and L1-to-L2 settings when compared to L1 Adaptation.

\paragraph{Specific adaptation transfers better to related categories across ontology.} 

Although the two data sources in \dataset differ considerably in style and content, their ontological categories partially overlap. For example, \emph{Mathematics} and \emph{Art} appear in both Wikipedia and Semantic Scholar. Is it possible to transfer between corresponding categories across ontologies? 

To answer this question, we first manually align L1 domains from Wikipedia and Semantic Scholar with similar ontological categories (e.g., grouping \emph{Mathematics} from Wikipedia and \emph{Mathematics} from S2ORC). We then apply a model adapted to an L1 domain in a source ontology onto its corresponding L1 domain in a target ontology. We compare this cross-ontology performance with two baselines: 1) the average out-of-domain performance of other L1 adapted models in the target ontology and 2) the in-domain performance of a model adapted to the target L1 domain. 

Our results are displayed in Table \ref{tab:cross_level}. We observe that while L1 adapted models are effective at transferring to other domains \emph{within} an ontology, they are less effective at transferring to corresponding domains \emph{outside} an ontology. Surprisingly, in all cases, transferring outside an ontology performs even worse than using the base GPT-2 model with no additional adaptation. Moreover, the average out-of-domain performance of L1 adapted models generally outperforms cross-ontology performance, indicating properties shared within an ontology (e.g. style) could be transferred.

\paragraph{Summary} Our investigations into the out-of-domain performance of adapted language models reveals a tradeoff between specialization and generalization. The more fine-grained the specialization of the language model, the less one can expect it to be applicable outside of the domain it was trained on. This effect size increases as we move outside the ontology: models trained on one ontology are not useful in other ontologies, despite being trained on similar categories of data. These findings lead us to believe that domain adaptation should be studied from a multi-faceted perspective to exploit specific aspects of domain (e.g. style, content). Future work may look at reducing the tradeoff between highly domain specialized models and out of domain performance, perhaps through ensembling or other approaches.

\begin{figure}[t]

\includegraphics[width=0.48\textwidth]{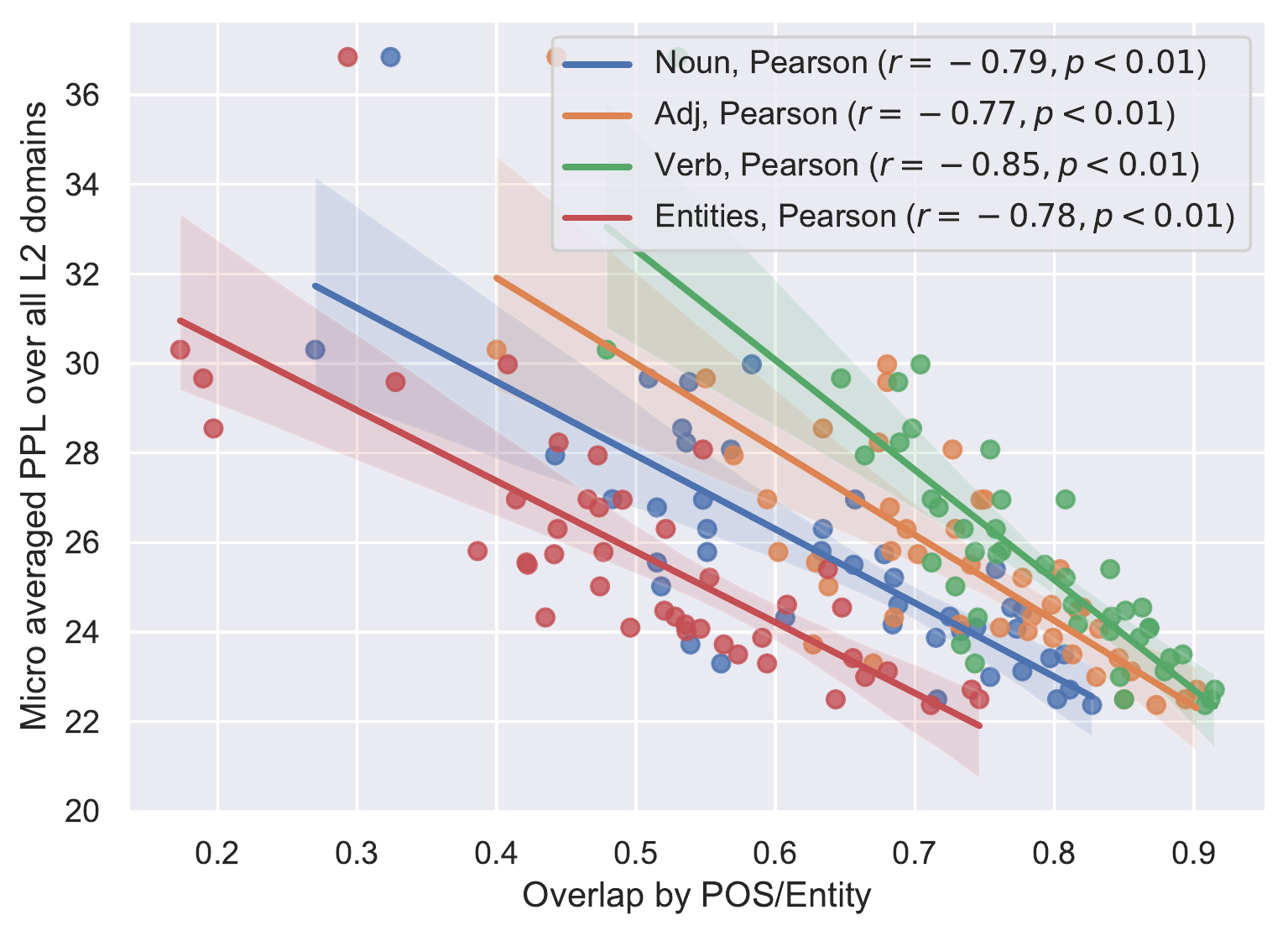}
\caption{The relationship between overlap and transfer performance over all Wikipedia L2 domains. Entities, verbs, nouns and adjectives are all strongly correlated with performance across domains.}
\label{fig:overlap}
\end{figure}
\subsection{Lexical indicators of out-of-domain performance}
\label{sec:lexical_features}

Looking closer at the out-of-domain performance of L1 models, we see intuitive relationships between subject similarity and zero-shot out-of-domain transfer performance (Table \ref{tab:transfer}). For example, \emph{Society} and \emph{Human Activities} domains tend to transfer well to each other, whereas \emph{Religion} and \emph{Mathematics} do not transfer as well. These findings suggest that out-of-domain transfer is correlated with content overlap.  In this section, we present some basic lexical indicators of out-of-domain performance which support this hypothesis.

% With the development of \dataset, we can now study features that may be important for transfer among domains. We look at some examples of properties that play a role in transfer between domains and explore their role in more depth. 
% Specifically, we compare vocabulary overlap for different parts of speech and entities, as well as analyze the shift in vocabulary distribution after adaptation.
\paragraph{Vocabulary overlap strongly correlates with transfer regardless of part-of-speech.}

% \suchin{For this figure, I would display it as a regression scatter (use `sns.regplot` in seaborn), and report the pearson correlation  (and p-value) }
Figure~\ref{fig:overlap} shows the correlation of vocabulary overlap a given part-of-speech tag (VERB, NOUN, ADJ) or entities and average out-of-domain performance on \dataset. 
We compute this by taking the top-$k$ ($k=1000$) most common words for a given domain which correspond to a given POS tag. For every given domain, we then calculate the intersection of shared most common words corresponding to the part-of-speech tag with the entirety of \dataset and plot them against the L2-domain-averaged perplexity over the entire dataset. We use \texttt{spacy} \citep{spacy2} for both entity recognition and POS tagging. We find that vocabulary overlap is a strong predictor of transfer performance regardless of part-of-speech, perhaps indicating its relevance in transfer between fine-grained domains.
\begin{table}[t]
    \centering
    \resizebox{0.48\textwidth}{!}{
    \begin{tabular}{ccccc}
    \toprule
         Transfer & Domain-specific & General & Examples  \\
    \midrule
        Distant L1&25.7\% & 74.3\%  & Blockchain, Alexa \\ 
         Easy L1 & 12.3\% & 87.7\% & the, cache\\ 
         Zero-shot & 23.4\% &  76.6\% & renewals, Markov \\
         L1-to-L2 & 31.6\% &  68.4\% &  lambda DCS, Tacotron\\
    \bottomrule
         
    \end{tabular}}
    \caption{Average percentage of tokens transferred in-domain and out of domain. Examples are taken from Philosophy$\rightarrow$Computer Science, Statistics$\rightarrow$Computer Science, GPT2$\rightarrow$Computer Science, and Computer Science$\rightarrow$Computation and Language.}
    \label{tab:dom_tokens}
\end{table}
% \begin{table}[]
%     \centering
%     \resizebox{0.48\textwidth}{!}{
%     \begin{tabular}{ccccc}
%     \toprule
%          Transfer & NOUN$_{\text{\%}\uparrow}$ & VERB$_{\text{\%}\uparrow}$ & PROPN$_{\text{\%}\uparrow}$ & ADJ${\text{\%}\uparrow}$  \\
%     \midrule
%         Distant L1& \\ 
%          Easy L1 &\\ 
%          Zero-shot & \\
%          L1-to-L2 &  \\
%     \bottomrule
         
%     \end{tabular}}
%     \caption{Distribution of most common POS tags after four types of transfer. }
%     \label{tab:pos_dist}
% \end{table}

\paragraph{Related domains mostly transfer domain-specific tokens.}
%  \mr{Analysis with domain specific tokens similar to deepmind paper that suchin shared}

% \suchin{I'm a little bit confused about this section; are you first adapting to a source domain, and then adapting to a target domain? (what is the "transfer pair"?) How does comparing the probability distributions of two adaptations tell you what's learned in one of them? I expected an analysis comparing the adapted probability distribution with an unadapted one; might be missing something here. Furthermore, I think there are a few things left to be desired in this analysis; is it possible to quantify your results more rigorously? you could take a list of stopwords and present the \% of time they show up in the most "adapted" words? Or do something similar with pronouns/nouns/entities? What is the percent improvement in these perplexities, based on whether the domain is "near" or "distant"?}
% \mr{Add tables with appropriate quantitative analyses as discussed in the text}
% \suchin{Leaving this last paragraph to when the results table/figure is up}

We analyse domain adaptation at a token-level to characterize what different adaptation settings transfer. Specifically, we measure which tokens are most impacted in terms of per-word perplexity when we finetune on a domain-specific corpus. We do this by taking the difference between the softmax-normalized probability of predicting a given word in a given domain when comparing two models adapted to different corpora.
 
We compare S2ORC adapted models in four settings: two best-transferred domains (a proxy for similar domains; easy transfers), two worst transferred L1 domains (a proxy for distant domains; difficult transfers), L1-to-L2 Adaptation (hierarchical domain transfer), and no adaptation (zero-shot performance of the base LM).
We show the distribution between domain-specific (terms that appear less than 0.00001\% of the time in any other domain) and non-domain-specific terms in Table~\ref{tab:dom_tokens} that appear in the top 1000 most adapted words.
Finally, we show representative samples of tokens with the greatest change after adaptation. 
We find that the most changed tokens between easy transfers (e.g. Statistics and Computer Science) are non-domain-specific words (such as \emph{the})
 %\suchin{how are you defining domain specific? eyeballing or something quantitative?}
but harder transfers include words that are more domain specific (such as \emph{Blockchain}).
%  \begin{table}[]
%      \centering
%      \footnotesize
%      \begin{tabular}{cc}
% 	     \toprule
% 		Example Transfer Pair& Samples from top 100 words\\
% 		\midrule
% 	     HUMA $\rightarrow$ RELI  & Allah, Synagogue, Jesus,  \\
% 	     MATH $\rightarrow$ RELI & God, \emph{the}, \emph{we} \\
% 	     \bottomrule
%      \end{tabular}
%      \caption{Transferred tokens between distant and similar domains (measured using original transfer perplexity).}
%      \label{tab:my_label}
%  \end{table}

% \subsection{Model scale}
% \suchin{This section could also be moved to after 3.3.}
% \mr{running experiments with 345M GPT2}
% \paragraph{How does pre-training data impact adaptation to specific domains?}
% \mr{SciGPT2 from AI2}
% \mr{pre-training on wikipedia vs scientific}
% \mr{probably push to future work}
\paragraph{Summary} Our preliminary analyses suggest that simple lexical characteristics of domains are strong indicators of how well an adapted model may generalize.  Developing computationally inexpensive indicators of transfer (as lexical overlap is), is important for domain transfer to find the best out of a large set of candidate corpora to perform adaptation to a target domain. This would allow one to approximately find the best corpus, without the computational overhead of adapting to all candidate corpora. % \suchin{what else do we want to say here?}

\section{Related Work}
\paragraph{Domain Adaptation Techniques}
\citep{gururangan2020dont} show that pretrained language models can be adapted to new domains by continued pre-training on domain-specific corpora. \citet{Chronopoulou,gururangan2021demix} build upon this work by using hierarchically constructed domain specific adapters/experts \citep{houlsby2019parameterefficient}. Another line of work in domain generalization is to simply scale the model pre-training on a corpus containing different domains (e.g.~GitHub, PubMed) such as done with GPT-J \citep{gpt-j} and the Pile \citep{gao2021pile}. \citet{dery2021should} also look to bridge these approaches by learning a task/domain specific mixture of tasks. Overall, however, much of this work \citep{daume-iii-2007-frustratingly,ruder2017knowledge,ruder-plank-2018-strong,gururangan2020dont,ramponi-plank-2020-neural,gururangan2021demix,Chronopoulou} fits in a paradigm in which a base model is trained further on domain-specific corpora and then testing on tasks within that domain (e.g.~abstract sentence role classification \citep{bird-etal-2008-acl} for the scientific domain).
\dataset is complementary to these works in providing a testbed for fine-grained and hierarhical adaptation across a large quantity of domains. 

% \paragraph{Domain Specific Models}
% With the rapid adoption of pre-trained language models by the NLP community, there has been a growing emphasis on including domain information during pre-training \citep{daume-iii-2007-frustratingly,gururangan2020dont}. To this end, many domain-specific pre-trained language models have been released that improve performance on in-domain datasets, such as SciBERT (scientific text; \citealp{beltagy2019scibert}), LegalBERT (legal text; \citealp{chalkidis2020legalbert}), FinBERT (financial BERT; \citealp{araci2019finbert}), and others. Although some models have been pre-trained from scratch given enough in-domain data, other domains are often fine-grained to the degree where adaptation is a more effective approach. \dataset can also be extended to include downstream tasks assigned to corresponding domains---this would allow for studies on the domain adaptation techniques with an eventual downstream component.

\paragraph{Domain Adaptation Datasets} One approach toward improved pre-trained language models includes building large-scale pre-training datasets that contain a diverse set of domains, such as the Pile \citep{gao2021pile}. Overall, this emphasis has lead to improved performance in various domains, especially with large-scale pre-trained language models, such as GPT-J \citep{gpt-j}. Another line of work has been in documenting large-scale web-crawled datasets, so practitioners and researchers can be more informed and mindful of the data used \citep{dodge-etal-2021-documenting}. Our work extends this thread with a massively multi-domain corpus with a manually curated ontology that can be used to study fine-grained and hierarchical domain transfer.

% However, despite this noted importance of training language models to be utilized in multiple domains, language models are not evaluated on fine-grained, massively multi-domain settings. 
\section{Conclusion}
We developed \dataset, a new massively multi-domain language modeling dataset for studying domain adaptation in language models. \dataset consists of 145 fine-grained domains (curated from Wikipedia and Semantic Scholar) that are hierarchically organized using domain-specific ontologies. Using \dataset, we find that domain precision is more important than data quantity to improve in-domain performance, a tradeoff between specialization and out-of-domain generalization. We release \dataset publicly to spur further research on building effective language models on highly heterogeneous data.

\section{Limitations}

In this work, we only consider adaptation techniques that assume domains are monolithic and non-overlapping. Future work may instead explore modeling the data as a mixture of domains, which may improve out-of-domain performance. In addition, \dataset only covers two data sources (Wikipedia and Semantic Scholar). Future work could expand this corpus with ontologies from other data sources, such as Reddit, which have a fine-grained and hierarchical domains. Moreover, data sourced from the web may contain hate speech and other harmful content, which may be reproduced by language models adapted to such data. The data sources we use adhere to research-friendly data licenses, but training models on web-curated data while maintaining the rights of authors as data subjects and creators remains an open problem.

\section*{Acknowledgements}
We thank Nikita Haduong, Jungo Kasai, Sophia Serrano, Wenya Wang for their feedback and proofreading comments. We also thank Sebastian Ruder for useful discussions. MR is grateful to the Masason Foundation for their support.
% \dataset is an important step towards the study of domain adaptation in NLP, however given that it tackles the broad notion of \textit{domain}, it consists of certain limitations. 

% Notable limitations include the lack of wide-spanning domain coverage comparable to that of webtext, a tradeoff when using manually curated ontologies. Furthermore, this dataset is developed to help study adaptation, in which language models could learn toxic biases from corpora during adapatation. Additionally, although study of adaptation could lead to increasingly efficient training of models for many specific use cases, it also may lead to better adaptation for nefarious reasons by bad actors. This being said, however, we believe that the \mr{benefits outweigh the risks?}
% \mr{
% - \dataset doesn't cover as wide-spanning coverage as webtext does \\
% - language models could learn bias from the sources in the corpus \\
% - adaptation can be studied and that can be a good thing, but it may lead to better adaptation of language models can be adapted for nefarious reasons}
%\suchin{we don't consider a mixture of domains}

\bibliography{custom}
\bibliographystyle{acl_natbib}

\onecolumn
\newpage
\appendix

\section{Appendix}
\label{sec:appendix}

\subsection{Hyperparameters}

\begin{table*}[!ht]
    \centering
    \small

    \begin{tabular}{cc}
      \toprule
      \textbf{Computing Infrastructure} & 8 Volta 16GB GPUs\\ 
      \bottomrule
    \end{tabular}
    
    \vspace{3mm}\begin{tabular}{cc}
        \toprule
        \textbf{Hyperparameter} & \textbf{Assignment}  \\
        \midrule
        architecture & GPT-2 \\
        \midrule
        tokens per sample & 1024 \\
        \midrule
        batch size & 64000 \\
        \midrule
        number of workers & 8 \\
        \midrule
        learning rate & 5e--5 \\
        \midrule
        clip norm & 0.1 \\
        \midrule
        number of steps & 1,000,000 \\
        \midrule
        save interval updates & 1,000 \\
        \midrule
        validation interval & 1,000 \\
        \midrule
        number of warmup steps & 10,000 \\
        \midrule
        learning rate scheduler & polynomial decay \\
        \midrule
        learning rate optimizer & Adam \\
        \midrule
        Adam beta weights & (0.9, 0.99) \\
        \midrule
        Adam epsilon & 1e-6 \\
        \midrule
        weight decay & 0.1 \\
        \bottomrule
    \end{tabular}
    
    \caption{Hyperparameters for finetuning in all settings.} 
    \label{tab:125M_hps}
\end{table*}

% \subsection{L2 Transfer Results}
% We release an interactive chart with all L2 transfer results as \texttt{L2\_s2orc.html} and \texttt{L2\_wikipedia.html} in our code relase. 
\subsection{Licenses}
Our data sources have open licenses. Wikipedia has a Creative Commons Attribution-ShareAlike 3.0 Unported License and a S2ORC has a Creative Commons Attribution-NonCommercial 4.0 International (CC BY-NC 4.0).
\subsection{More examples of most transferred tokens}
We give more examples of tokens transferred from the L1 S2ORC Computer Science (given its assumed familiarity to our audience) domain in the following table:
\begin{table*}[!htb]
    \centering
    \begin{tabular}{p{0.5\textwidth}p{0.5\textwidth}}
    \toprule
        \bf Transfer & \bf Example Tokens \\
        \midrule
        Computer Science$\rightarrow$Computation and Language & lambda DCS, perplexity, Artetxe, Tacotron, Swayamdipta, Transformer, parallel, Socher, Gigaword, Lapata  \\
        Computer Science$\rightarrow$Machine Learning &  criterion, Ganchev, Ioffe, labeling, autoencoder, Hinton, hyperparameters \\
        Computer Science$\rightarrow$Art & Atheist, heroism, intellectuals, horrors, witchcraft, mourning, apostles \\ 
        Computer Science$\rightarrow$Technology and Applied Sciences & Sunderland, accounting, inventory, Libyan, bishop, ravaged, traffic \\
        \bottomrule
    \end{tabular}
    \caption{More examples of most transferred tokens}
    \label{tab:most_transferred_tokens}
\end{table*}
\subsection{All domains}
We list all domains contained within dataset in Table~\ref{tab:all domains}.
\begin{table}[!ht]
    \centering
    \begin{tabular}{p{\linewidth}}
        \toprule
        \bf S2ORC \\
        \midrule
cs.CE, cs.IT, cs.CG, cs.SI, cond-mat.quant-gas, math.SG, cs.SC, cs.CY, econ.GN, math.CO, cs.AR, cs.MS, cs.DC, q-bio.TO, cs.GR, physics.acc-ph, physics.geo-ph, math.RT, math.HO, cs.RO, q-bio.SC, math.QA, cs.NI, math.CA, cs.DS, astro-ph.GA, physics.atom-ph, math.CT, cs.CV, cond-mat.mtrl-sci, math.CV, math.AC, cond-mat.str-el, physics.comp-ph, cs.CC, math.FA, cond-mat.dis-nn, econ.TH, physics.gen-ph, physics.data-an, astro-ph.IM, q-bio.CB, math.LO, physics.ins-det, q-bio.BM, cs.LO, math.GR, physics.optics, cs.GT, math.AG, cs.NE, cs.SY, physics.bio-ph, physics.flu-dyn, cs.CL, math.MG, cs.AI, math.OC, nlin.CG, math.IT, stat.OT, math.OA, cond-mat.soft, Art, cs.GL, cs.PF, math.ST, physics.ao-ph, physics.plasm-ph, math.RA, physics.hist-ph, cs.PL, cs.MA, physics.chem-ph, physics.soc-ph, physics.med-ph, physics.ed-ph, stat.AP, stat.CO, math.DS, cs.DB, nlin.SI, q-bio.GN, physics.atm-clus, nlin.CD, astro-ph.CO, cs.CR, cond-mat.supr-con, cs.LG, math.KT, stat.ML, nlin.PS, q-bio.MN, cs.IR, math.GT, cs.SD, math.NA, cond-mat.other, math.NT, cs.FL, physics.pop-ph, cond-mat.stat-mech, math.GN, cs.DL, astro-ph.EP, q-bio.QM, cs.ET, q-bio.PE, cs.OH, Philosophy, physics.space-ph, econ.EM, physics.class-ph, cs.DM, cond-mat.mes-hall, stat.TH, cs.SE, astro-ph.HE, math.MP, nlin.AO, math.AP, q-bio.NC, q-bio.OT, astro-ph.SR, math.DG, math.AT, cs.MM, stat.ME, cs.OS, math.SP, physics.app-ph, cs.NA, math.PR, math.GM, cs.HC \\
         \midrule\\
         \toprule
         \bf Wikipedia \\
         \midrule
         Culture and Humanities, Games and Toys, Mass media, Performing arts, Sports and Recreation, The arts and Entertainment, Visual arts, Further research tools and topics, Reference works, Exercise, Health science, Human medicine, Nutrition, Public health, Self care, By continent, By period, By region, Human activities, Impact of human activity, Fields of mathematics, Logic, Mathematics, Biology, Earth sciences, Nature, Physical sciences, Philosophy, Thinking, Allah, Belief systems, Major beliefs of the world, Social sciences, Society, Agriculture, Computing, Engineering, Transport\\
         \bottomrule
    \end{tabular}
    \caption{All domains contained within \dataset}
    \label{tab:all domains}
\end{table}
\end{document}